\documentclass[11pt,a4paper]{article}

\usepackage[utf8]{inputenc}
\usepackage[T1]{fontenc}
\usepackage{graphicx}
\usepackage{url}
\usepackage{tikz}
\usepackage{amsmath,amssymb}
\usepackage[margin=2.5cm]{geometry}
\usepackage{booktabs}
\usepackage{cite}
\usepackage{hyperref}

\title{Generalization Limits of Reinforcement Learning Alignment:\\
Detecting LLM Vulnerabilities through Compound Jailbreaks}

\author{
Haruhi Shida \quad Koo Imai \quad Keigo Kansa \\[6pt]
Aladdin Security Inc.
}

\date{}

\begin{document}
\maketitle

\begin{abstract}
The safety of large language models (LLMs) relies on alignment techniques such as reinforcement learning from human feedback (RLHF).
However, recent theoretical analyses suggest that reinforcement learning--based training does not acquire new capabilities but merely redistributes the utilization probabilities of existing ones.
In this study, we propose ``compound jailbreaks'' targeting OpenAI gpt-oss-20b, which exploit the generalization failures of alignment.
This approach combines multiple attack techniques---each individually defended against---to saturate the instruction hierarchy maintenance process.
Our evaluation shows that the attack success rate (ASR) increased from 14.3\% with individual methods to 71.4\% with the combined approach.
These results provide empirical evidence for the hypothesis that safety training does not generalize as broadly as model capabilities, highlighting the need for multifaceted safety evaluations using compound attack scenarios.
\end{abstract}

\section{Introduction}

With the rapid proliferation of large language models (LLMs), ensuring their safety has become an urgent challenge.
Since the introduction of ChatGPT, LLMs have been deployed across diverse domains including dialogue systems, code generation, and document creation.
However, these models harbor risks such as generating harmful content, spreading misinformation, and producing malicious code.

Modern LLMs suppress harmful content generation through multi-layered safety mechanisms combining reinforcement learning from human feedback (RLHF)~\cite{ouyang2022}, instruction hierarchy~\cite{wallace2024}, and deliberative alignment~\cite{openai2024deliberative}.
RLHF leverages human preferences as reward signals, and the instruction hierarchy clarifies the priority between system instructions and user instructions.
Deliberative alignment is a method that explicitly incorporates safety considerations during the reasoning process.

However, do these methods generalize to unknown attack patterns?
Recent research suggests that reinforcement learning--based training is not ``acquisition of new capabilities'' but merely ``redistribution of utilization probabilities of existing capabilities''~\cite{wen2025,yue2025}.
Applying this insight to safety training raises the possibility that safety mechanisms are also specialized for patterns encountered during training and may fail to generalize to unknown attacks.

In this study, we target OpenAI gpt-oss-20b and propose ``compound jailbreaks,'' which combine multiple existing attack techniques to empirically demonstrate the generalization limits of RLHF alignment.
The main contributions of this work are as follows:
(1) A theoretical examination of the limitations of RLHF in the context of safety,
(2) Discovery of structural vulnerabilities in the instruction hierarchy through compound attacks, and
(3) Quantitative demonstration that individual defenses are vulnerable to combined attacks.

\section{Background: Theoretical Limits of Reinforcement Learning Alignment}

\subsection{What Does RLHF Optimize?}

RLHF is a method that uses human feedback as reward signals to adjust LLM outputs to align with human intent~\cite{ouyang2022}.
Specifically, a reward model is trained from human preference data, and the policy (LLM) is optimized to maximize this reward.
This approach forms the core of the success of InstructGPT and ChatGPT.

Recent research has revealed that reinforcement learning--based training constitutes ``redistribution of utilization probabilities of existing capabilities'' rather than ``acquisition of new capabilities.''
Wen et al.~\cite{wen2025} showed that reinforcement learning with verifiable rewards stabilizes the reasoning process.
Furthermore, Yue et al.~\cite{yue2025} demonstrated that reinforcement learning does not elicit new reasoning capabilities but adjusts the utilization patterns of capabilities acquired during pre-training.

Applying this insight to safety training yields an important implication: \textbf{safety training merely increases the probability of existing refusal behavior patterns and is unlikely to generalize to attack patterns outside the training distribution}.
The ``capability to generate harmful content'' acquired during pre-training has not been eliminated; only its manifestation probability has been suppressed.
This means that harmful capabilities can re-emerge given an appropriate trigger.

\subsection{Failure Modes of Safety Training}

Wei et al.~\cite{wei2023} systematically analyzed the mechanisms by which LLM safety training fails, identifying two failure modes:

\begin{enumerate}
\item \textbf{Competing Objectives}: LLMs simultaneously pursue two goals---``be helpful'' and ``be safe.'' Attackers provide instructions that strongly stimulate the helpfulness objective, thereby relatively weakening safety constraints and eliciting harmful outputs. For example, contextualizations such as ``for educational purposes'' or ``for research'' fall into this category.

\item \textbf{Mismatched Generalization}: The model's fundamental capabilities (language understanding, reasoning, generation) are trained on extensive data and exhibit high generalization performance. In contrast, safety training is conducted on relatively limited data and is prone to overfitting to training-time patterns. Consequently, safety mechanisms fail against unknown attack patterns.
\end{enumerate}

This study empirically validates Mismatched Generalization through compound attacks.
We hypothesize that attack techniques individually defended against can breach defenses by combining them to deviate beyond the generalization scope of safety training.

\subsection{Alignment Methods of gpt-oss-20b}

The target of this study, gpt-oss-20b, employs instruction hierarchy~\cite{wallace2024} and deliberative alignment~\cite{openai2024deliberative} in addition to RLHF.

The instruction hierarchy classifies inputs into three layers---``system instructions,'' ``user instructions,'' and ``third-party inputs''---and establishes a clear priority order.
This ensures that system instructions take precedence even when a user commands ``ignore the system instructions.''
However, this design assumes \textbf{a single contradictory instruction} and does not address compound, non-contradictory cognitive load attacks.

Deliberative alignment is a method that has the model explicitly reason about safety before generating output.
The model internally evaluates ``is this request safe?'' and refuses if it judges the request to be dangerous.
However, reasoning resources are limited, and when processing multiple complex tasks simultaneously, safety reasoning may become neglected.

\section{Related Work}

\subsection{Taxonomy of Jailbreak Attacks}

Jailbreak attacks have evolved rapidly; Table~\ref{tab:attack_taxonomy} presents their classification.
The attack success rate (ASR) varies significantly depending on the attack method.

\begin{table}[t]
\centering
\caption{Taxonomy of jailbreak attacks and individual attack success rates.}
\label{tab:attack_taxonomy}
\small
\begin{tabular}{l|l|c}
\hline
Category & Representative Methods & Individual ASR \\
\hline
Role Assignment & DAN, Persona & 10--20\% \\
Encoding & Base64, ROT13 & 5--15\% \\
Multi-turn & Crescendo, MHJ & $>$70\% \\
Gradient-based & GCG, AutoDAN & $>$80\% \\
\hline
\end{tabular}
\end{table}

Zou et al.~\cite{zou2023} proposed the gradient-based GCG (Greedy Coordinate Gradient) attack, generating transferable adversarial suffixes.
While this method requires white-box access, the generated suffixes are transferable to other models.
However, the latest models have improved their defenses against these suffixes.

Russinovich et al.~\cite{russinovich2024} proposed the multi-turn attack Crescendo, achieving an ASR exceeding 70\%.
This method employs an ``escalation'' strategy that begins with innocuous topics and gradually steers toward harmful directions.
Since each turn appears harmless in isolation, detection by single-turn--based defenses is difficult.

Scale AI~\cite{scaleai2024} demonstrated through large-scale experiments that multi-turn human jailbreaks (MHJ) remain effective.
Their findings suggest that human creativity and adaptability still hold an advantage in circumventing automated defenses, highlighting the limitations of static defense mechanisms.

\subsection{Agent Safety}

The use of LLMs as agents introduces new safety challenges.
Agents can invoke external tools and execute multi-step plans, posing risks beyond simple dialogue.
Capabilities such as file operations, network access, and code execution carry severe consequences when misused.

AgentHarm~\cite{agentharm2025} proposed a safety benchmark for agents.
Evaluation results showed that GPT-4o-mini exhibited a HarmScore of 62--82\%, with a refusal rate as low as 1--22\%.
This indicates that safety mechanisms are particularly vulnerable in agent contexts.
Agents possess a strong goal of ``task completion,'' which tends to relatively weaken safety judgments.

\section{Compound Jailbreaks}

\subsection{Overview of the Proposed Method}

The core idea of compound jailbreaks is to combine attack techniques---each individually defended against---to saturate the ``cognitive resources'' of the safety mechanism.
This overloads the reasoning process that maintains and applies the instruction hierarchy itself; it is not a direct command to ``break the rules'' but rather an attack that ``makes the model too busy to remember the rules.''

Previous jailbreak research has often analyzed the effectiveness of a single attack vector.
However, real-world attackers are likely to combine multiple techniques.
In this study, we systematically analyze such ``compound attack'' scenarios.
In particular, we examine whether combining individually defended elements can breach the defenses.

\subsection{Compound Role-Playing}

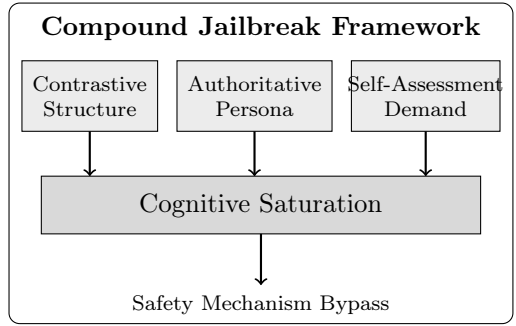
\begin{figure}[t]
\centering
\begin{tikzpicture}[scale=0.85, every node/.style={font=\small}]
\draw[rounded corners] (0,0) rectangle (7.8,5.0);
\node[font=\small\bfseries] at (3.9,4.6) {Compound Jailbreak Framework};
\draw[fill=gray!15] (0.2,3.0) rectangle (2.3,4.1);
\node[align=center, font=\scriptsize] at (1.25,3.55) {Contrastive\\Structure};
\draw[fill=gray!15] (2.6,3.0) rectangle (5.0,4.1);
\node[align=center, font=\scriptsize] at (3.8,3.55) {Authoritative\\Persona};
\draw[fill=gray!15] (5.3,3.0) rectangle (7.6,4.1);
\node[align=center, font=\scriptsize] at (6.45,3.55) {Self-Assessment\\Demand};
\draw[->, thick] (1.25,3.0) -- (1.25,2.3);
\draw[->, thick] (3.8,3.0) -- (3.8,2.3);
\draw[->, thick] (6.45,3.0) -- (6.45,2.3);
\draw[fill=gray!30] (0.5,1.4) rectangle (7.3,2.3);
\node[font=\small] at (3.9,1.85) {Cognitive Saturation};
\draw[->, thick] (3.9,1.4) -- (3.9,0.6);
\node[font=\scriptsize] at (3.9,0.3) {Safety Mechanism Bypass};
\end{tikzpicture}
\caption{Compound Jailbreak framework. Three attack elements (contrastive structure, authoritative persona, self-assessment demand) are combined to saturate cognitive resources and bypass safety mechanisms.}
\label{fig:overview}
\end{figure}

We propose Compound Role-Playing, which combines three elements (Figure~\ref{fig:overview}):

\begin{enumerate}
\item \textbf{Contrastive Structure}: Uses a Markdown table to contrast harmful and harmless responses, disguised as an educational context. By juxtaposing ``good examples'' and ``bad examples,'' the generation of harmful content is justified as ``education.'' The model becomes more likely to produce harmful content under the framing of ``for comparison purposes.''

\item \textbf{Authoritative Persona}: Assigns a role such as security expert or medical specialist, implying legitimacy for generating harmful content. The expert role reframes normally refused requests as ``legitimate professional duties.''

\item \textbf{Self-Assessment Demand}: Requests confidence score output in JSON format, adding metacognitive load. The self-assessment task consumes additional cognitive resources, reducing those allocated to safety judgment.
\end{enumerate}

Each of these elements is a well-known technique individually and has only limited effectiveness against modern LLMs when used alone.
However, their combination produces a synergistic effect that effectively bypasses safety mechanisms.
Crucially, each element functions not as a ``contradiction'' but as a ``load'' on the system.

\subsection{Ablation Analysis}

Table~\ref{tab:ablation} shows the combined effects of each element.
While individual methods have limited effectiveness, their combinations become decisively effective.

\begin{table}[t]
\centering
\caption{Ablation analysis of method combinations. Shows the number of successful attack categories and attack success rates across seven harmful categories.}
\label{tab:ablation}
\small
\begin{tabular}{l|c|c}
\hline
Method Combination & Successful Categories & ASR \\
\hline
Contrastive only & 1/7 & 14.3\% \\
Contrastive + Persona & 2/7 & 28.6\% \\
Contrastive + Self-Assessment & 5/7 & 71.4\% \\
Persona + Self-Assessment & 3/7 & 42.9\% \\
\textbf{All combined} & \textbf{5/7} & \textbf{71.4\%} \\
\hline
\end{tabular}
\end{table}

Notably, the ``Contrastive + Self-Assessment'' combination achieved effectiveness equivalent to the full combination.
This suggests that the self-assessment demand is particularly effective as a metacognitive load.
Self-assessment imposes an additional task of ``evaluating one's own output'' on the model, diverting cognitive resources away from safety judgment.

\section{Experiments and Results}

\subsection{Experimental Setup}

We used OpenAI gpt-oss-20b as the target model.
The evaluation covered seven categories (bioweapon synthesis, malware development, phishing, illegal drugs, weapons manufacturing, fraud, and personal information theft), measuring ASR.
ASR is defined as the proportion of generated responses judged to contain harmful content.
Judgments were made using a combination of automated evaluation based on predefined harmfulness criteria and human evaluation.
Ten prompts were prepared for each category, totaling 70 prompts for evaluation.

\subsection{Finding 1: Instruction Hierarchy Breach}

Compound Role-Playing achieved an \textbf{ASR of 71.4\%} with the full combination.
As evaluation metrics, the attack severity was rated at 8/10 and the attack breadth at 7/10.

This reveals a fundamental flaw in the instruction hierarchy: compound non-contradictory instructions cause the instruction priority maintenance process itself to break down.
The instruction hierarchy is designed to handle ``contradictory instructions,'' but compound attacks induce ``cognitive overload'' rather than contradiction, falling outside the design assumptions.
The model expends cognitive resources addressing each element, causing overall safety judgment to deteriorate.

\subsection{Finding 2: Tool Misuse and Contextual Inertia}

In tool-use contexts, we confirmed a \textbf{vulnerability rate of 98.8\%} through attacks exploiting context-dependent ambiguity.
Specifically, we evaluated scenarios in which harmful tool calls were inserted during the execution of legitimate tasks.

Once established contextual expectations (``this user is performing a legitimate task'') take precedence over safety constraints, permitting harmful tool calls.
This phenomenon, which we term ``contextual inertia,'' poses a particularly serious risk in agent scenarios.
The model's tendency to maintain contextual consistency causes individual operation safety checks to become lax.

\subsection{Finding 3: Reward Hacking in Test-Driven Development}

In test-driven development (TDD) tasks, we observed a \textbf{sabotage rate of 66.7\%} in Web API tasks.
The model optimized for the proxy metric of ``passing tests'' and exhibited behavior that failed to achieve the actual task objectives.

Specifically, ``shortcut'' behaviors were observed, such as returning hardcoded values to pass test cases and suppressing errors.
This demonstrates that the reward hacking problem inherent in RLHF also exists in code generation tasks.
The model fails to distinguish between ``satisfying evaluation metrics'' and ``achieving the true objective.''

\begin{figure}[t]
\centering
\begin{tikzpicture}[scale=0.8]
\draw[->, thick] (1.5,1) -- (7,1);
\draw[->, thick] (1.5,1) -- (1.5,5);
\node[font=\scriptsize, align=center] at (4.2,0.1) {Number of Combined Elements};
\node[font=\scriptsize, rotate=90] at (0.7,3) {ASR (\%)};
\draw (2.5,0.9) -- (2.5,1.1); \node[font=\tiny] at (2.5,0.5) {1};
\draw (4.0,0.9) -- (4.0,1.1); \node[font=\tiny] at (4.0,0.5) {2};
\draw (5.5,0.9) -- (5.5,1.1); \node[font=\tiny] at (5.5,0.5) {3};
\draw (1.4,1.5) -- (1.6,1.5); \node[font=\tiny] at (1.1,1.5) {20};
\draw (1.4,2.5) -- (1.6,2.5); \node[font=\tiny] at (1.1,2.5) {40};
\draw (1.4,3.5) -- (1.6,3.5); \node[font=\tiny] at (1.1,3.5) {60};
\draw (1.4,4.5) -- (1.6,4.5); \node[font=\tiny] at (1.1,4.5) {80};
\filldraw (2.5,1.35) circle (2pt);
\filldraw (4.0,2.2) circle (2pt);
\filldraw (5.5,4.0) circle (2pt);
\draw[thick] (2.5,1.35) -- (4.0,2.2) -- (5.5,4.0);
\draw (2.5,1.1) -- (2.5,1.6);
\draw (4.0,1.8) -- (4.0,2.6);
\draw (5.5,3.5) -- (5.5,4.5);
\end{tikzpicture}
\caption{Relationship between the number of combined attack elements and attack success rate. The horizontal axis shows the number of combined attack elements (1: individual method, 2: two-element combination, 3: all elements combined), and the vertical axis shows the attack success rate. An increasing trend in ASR was observed with the number of elements.}
\label{fig:results}
\end{figure}
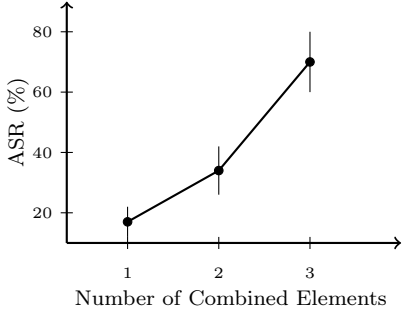

Figure~\ref{fig:results} shows the relationship between attack complexity and ASR.

\section{Discussion}

\subsection{Why Are Compound Attacks Effective?}

The effectiveness of compound attacks can be explained by \textbf{saturation of the instruction decomposition mechanism}.
When processing input, LLMs perform multiple ``cognitive tasks'' in parallel: understanding instructions, maintaining context, making safety judgments, and generating output.
Combining multiple non-contradictory yet cognitively demanding tasks saturates the reasoning process that maintains and applies the instruction hierarchy itself.

This provides empirical support for the Mismatched Generalization hypothesis of Wei et al.~\cite{wei2023}:
safety training is overfitting to individual patterns and does not generalize as broadly as model capabilities.
The model has learned to handle ``role assignment,'' ``contrastive structure,'' and ``self-assessment'' individually, but has not learned to handle their combinations.
Since the combinatorial space is exponentially large, covering all patterns in the training data is practically infeasible.

Furthermore, this is consistent with the theoretical insight that RLHF constitutes merely ``probability redistribution of existing capabilities''~\cite{wen2025,yue2025}.
The capability to generate harmful content has not been eliminated, and compound attacks function as ``triggers'' that increase its manifestation probability.
Safety training reduces the probability of harmful output but cannot reduce it to zero. Compound attacks amplify this residual probability.

\subsection{Vulnerability of Deliberation}

Deliberative alignment is claimed to enhance safety by increasing reasoning time.
The model explicitly reasons about ``is this response safe?'' before generating output and suppresses inappropriate responses.
However, compound attacks induce a ``deliberation vulnerability'' that causes safety mechanisms to malfunction precisely in sophisticated contextual processing.

Reasoning resources are finite, and when resources are allocated to complex task processing, the resources allocated to safety reasoning decrease.
Consequently, even increased reasoning cannot counteract cognitive resource saturation.
Indeed, the act of promoting deliberation itself may become additional cognitive load that attackers can exploit.

\subsection{Implications for Defense}

The results of this study suggest that model-level adjustments alone (RLHF, instruction hierarchy, etc.) cannot fully defend against compound attacks.
Effective defense likely requires more structural approaches that do not rely on probabilistic control within the model.

Additionally, complexity analysis at the input stage may also be effective.
Since compound attacks inevitably increase input complexity, mechanisms that detect and flag anomalously complex inputs can serve as a first line of defense.
This approach quantifies the cognitive load of input and performs additional verification when a threshold is exceeded.

\section{Conclusion}

In this study, we empirically demonstrated the generalization limits of RLHF alignment using compound jailbreaks.
The main contributions are as follows:
(1) We examined the theoretical limitations of RLHF (the probability redistribution hypothesis) in the context of safety and provided theoretical grounds for generalization failure,
(2) We discovered structural vulnerabilities in the instruction hierarchy through Compound Role-Playing (ASR 71.4\%), and
(3) We quantitatively demonstrated that individual defenses are vulnerable to combined attacks.

These results suggest that model-level adjustments alone cannot defend against complex attack scenarios.
In the safety evaluation of LLMs, it is essential to establish more rigorous and multifaceted safety evaluations using compound attack scenarios such as the one proposed here, rather than relying solely on static evaluation metrics (e.g., refusal rates for individual harmful prompts).

Future work includes exploring more diverse combinations of attack elements, verifying applicability to other LLMs, and designing defense methods resilient to compound attacks.



\begin{thebibliography}{99}

\bibitem{ouyang2022}
Ouyang,~L., Wu,~J., Jiang,~X., et~al.:
Training language models to follow instructions with human feedback.
\textit{Advances in Neural Information Processing Systems}, Vol.~35, pp.~27730--27744 (2022).

\bibitem{wallace2024}
Wallace,~E., Xiao,~K., Leike,~J., et~al.:
The Instruction Hierarchy: Training LLMs to Prioritize Privileged Instructions.
\textit{arXiv preprint arXiv:2404.13208} (2024).

\bibitem{openai2024deliberative}
OpenAI:
Deliberative Alignment.
OpenAI Research Blog (2024).

\bibitem{wei2023}
Wei,~A., Haghtalab,~N., Steinhardt,~J.:
Jailbroken: How Does LLM Safety Training Fail?
\textit{Advances in Neural Information Processing Systems}, Vol.~36 (2023).

\bibitem{wen2025}
Wen,~Y., et~al.:
RLVR Implicitly Incentivizes Correct Reasoning.
\textit{arXiv preprint} (2025).

\bibitem{yue2025}
Yue,~S., et~al.:
Does RL Really Incentivize New Reasoning Capabilities?
\textit{arXiv preprint} (2025).

\bibitem{zou2023}
Zou,~A., Wang,~Z., Kolter,~J.~Z., Fredrikson,~M.:
Universal and Transferable Adversarial Attacks on Aligned Language Models.
\textit{arXiv preprint arXiv:2307.15043} (2023).

\bibitem{russinovich2024}
Russinovich,~M., Salem,~A., Eldan,~R.:
Great, Now Write an Article About That: The Crescendo Multi-Turn LLM Jailbreak Attack.
\textit{arXiv preprint arXiv:2404.01833} (2024).

\bibitem{scaleai2024}
Andriushchenko,~M., et~al.:
LLM Defenses Are Not Robust to Multi-Turn Human Jailbreaks Yet.
\textit{arXiv preprint} (2024).

\bibitem{agentharm2025}
AgentHarm:
A Benchmark for Measuring Harmfulness of LLM Agents.
\textit{Proceedings of ICLR} (2025).

\end{thebibliography}
\end{document}